\pdfoutput=1%For ARXIV

\documentclass[5p]{elsarticle} %双栏
\usepackage{amssymb}
\usepackage{lineno}

\usepackage{multirow}
\usepackage{multicol}
\usepackage{booktabs}

\usepackage{subfigure}
\usepackage[ruled]{algorithm2e}
\usepackage{bbm}
\usepackage{amsmath} 
\usepackage[normalem]{ulem} % use normalem to protect \emph
\usepackage{makecell}

\newcommand{\slfrac}[2]{\left.#1\middle/#2\right.}
\newcommand\node[1]{\boldsymbol{v}_{#1}}
\newcommand\aVector[2]{\boldsymbol{#1}_{#2}}
\newcommand{\para}[1]{\vspace{2mm} \noindent \textbf{#1}}
\newcommand{\dis}{\mathcal{D}}

\journal{NeuroComputing}

\begin{document}
%\linenumbers

\begin{frontmatter}



\title{Pseudo Contrastive Learning for Graph-based Semi-supervised Learning}


\author[address1]{Weigang Lu\fnref{label1}}
\ead{wglu@stu.xidian.edu.cn}
\cortext[cor1]{Corresponding author}%corresponding author

\author{Ziyu Guan\fnref{label1}\corref{cor1}}
\ead{zyguan@xidian.edu.cn}

\author{Wei Zhao\fnref{label1}}
\ead{ywzhao@mail.xidian.edu.cn}

\author{Yaming Yang\fnref{label1}}
\ead{yym@xidian.edu.cn}

\author{Yuanhai Lv\fnref{label2}}
\ead{lvyuanhai@stumail.nwu.edu.cn}

\author{Lining Xing\fnref{label3}}
\ead{xinglining@gmail.com}

\author{Baosheng Yu\fnref{label4}}
\ead{baosheng.yu@sydney.edu.au}

\author{Dacheng Tao\fnref{label4}}
\ead{dacheng.tao@sydney.edu.au}

\address[label1]{School of Computer Science and Technology, Xidian University, Xi'an, China}
\address[label2]{School of Information Science and Technology, Northwest University, and the Information Network Center, Xi'an University of Posts and Telecommunications, Xi'an, China}
\address[label3]{School of Electronic Engineering, Xidian University, Xi'an, China}
\address[label4]{The University of Sydney, Australia}

\begin{abstract}
Pseudo Labeling is a technique used to improve the performance of semi-supervised Graph Neural Networks (GNNs) by generating additional pseudo-labels based on confident predictions. However, the quality of generated pseudo-labels has been a longstanding concern due to the sensitivity of the classification objective with respect to the given labels. To avoid the untrustworthy classification supervision indicating ``a node belongs to a specific class,'' we favor the fault-tolerant contrasting supervision demonstrating ``two nodes do not belong to the same class.'' Thus, the problem of generating high-quality pseudo-labels is then transformed into a relaxed version, i.e., identifying reliable negative pairs. To achieve this, we propose a general framework for GNNs, termed Pseudo Contrastive Learning (PCL). It separates two nodes whose positive and negative pseudo-labels target the same class. To incorporate topological knowledge into learning, we devise a topologically weighted contrastive loss that spends more effort separating negative pairs with smaller topological distances. Experimentally, we apply PCL to various GNNs, which consistently outperform their counterparts using other popular general techniques on five real-world graphs.
\end{abstract}


\end{frontmatter}

\section{Introduction}
\label{intro}

Over the past decades, deep learning has achieved great success in many fields, largely due to the extensive efforts of human annotators in labeling large amounts of training samples. However, unlike other data (e.g., pictures, languages, and videos), annotating graph data usually requires expertise in understanding not only the important properties of each node but also the complex relationships between different nodes. For example, identifying the category that a research paper belongs to often requires expensive efforts from the experts since many research works are associated with multiple different domains. Therefore, graph-based semi-supervised learning (SSL)~\cite{zhu2003semi,zhou2004learning,weston2012deep,yang2016revisiting}, which addresses heavy dependence on labeled data by leveraging the potential of unlabeled data, has recently attracted more attention from the research community. The motivation of graph-based SSL is to leverage the vast amount of unlabeled nodes, together with limited labeled nodes, to learn a model that can accurately predict the properties of graphs, such as the classes of nodes. In this paper, we focus on the \textbf{semi-supervised node classification} problem in graph-based SSL, which is one of the research highlights.

Graph Neural Networks (GNNs)~\cite{gcn,gat,gin,sgc,graphsage,gcnii,appnp,gprgnn} have emerged as the dominant technology in graph-based SSL. The success of GNNs can be attributed to their ability to exchange information between connected nodes. However, this mechanism limits the access of GNNs to only a small number of neighbor nodes in the local area defined by the model depth. As a result, GNNs struggle to effectively utilize the vast amount of unlabeled nodes beyond this confined area. Considering the different strengths of various GNNs (simplicity~\cite{sgc}, explainability~\cite{gat,ie-hgcn}, expressive ability~\cite{gin}, etc.), it motivates us to develop a general technique for GNNs to compensate for this shortcoming, i.e., \textit{limited ability to leverage unlabeled nodes}.

% PL to solve the problem in GNNs
Exploring extra labels is a typical answer to this problem. Pseudo Labeling (PL)~\cite{pl}, as a simple yet effective SSL approach, has been extensively used to generate additional pseudo-labels. We roughly categorize PL variants into two groups according to how pseudo-labels are generated and used: (1) \textit{Positive Pseudo Labeling (PPL)}~\cite{li2018deeper,sun2020multi,mutual_pl,dynamic_pl,informative_pl,shgp-yang}. It assigns positive pseudo-labels to unlabeled samples according to the maximum model prediction. Then, it teaches the model by Positive Learning (standard classification objective) to learn ``the samples should belong to these positive pseudo-labels.'' (2) \textit{Negative Pseudo Labeling (NPL)}~\cite{defense}. 
\begin{figure}[!ht]
     \centering
     \includegraphics[width=0.9\linewidth]{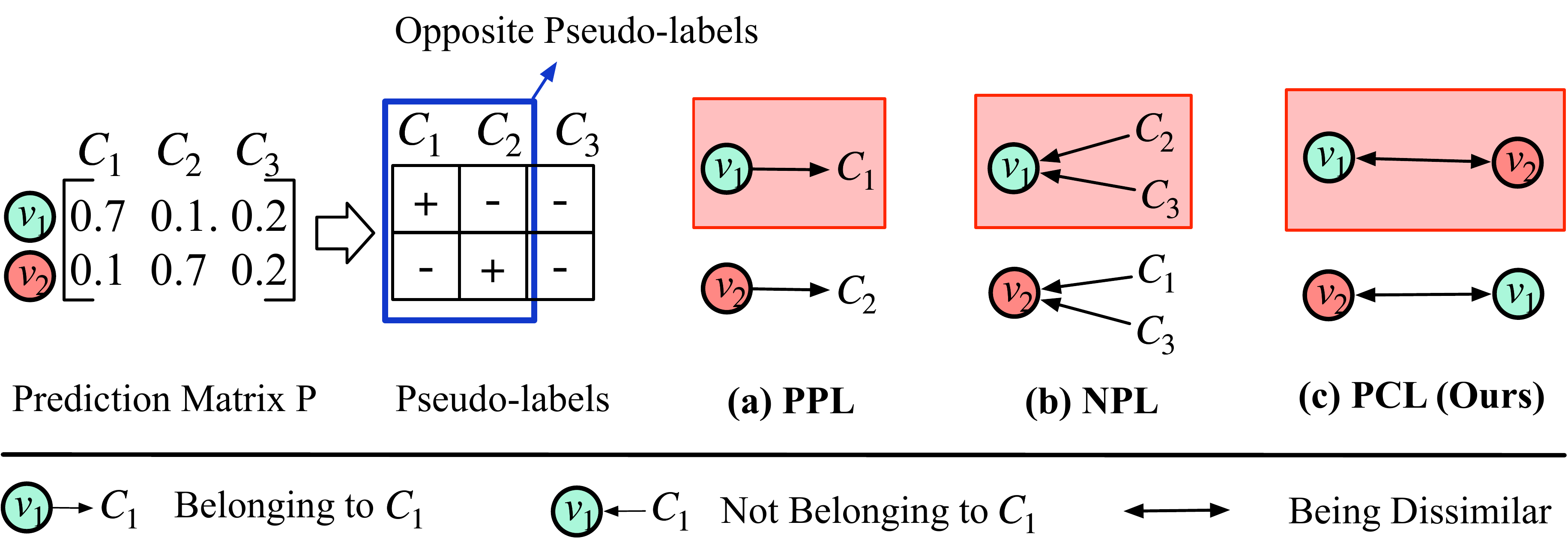}
     \caption{Comparison to PPL, NPL, and the combination of PPL and NPL.}
     \label{fig:PL_and_PCL}
\end{figure}For those unlabeled samples with sufficiently low predicted scores, it generates negative pseudo-labels and tells the model through Negative Learning~\cite{nlnl} that ``the samples should not belong to these negative pseudo-labels.''

% weakness of PPL and NPL
Compared to PPL which provides ``what to learn'' information, NPL which provides ``what not to learn'' information can decrease the risk of learning from incorrect information. It is due to the fact that there is only (possibly) one truth and numerous false instructions. PL-based\footnote{In the next content, we refer to PL/pseudo-labels as a general term for PPL and NPL/positive and negative pseudo-labels.} methods can exploit additional supervision, thereby benefiting the training process. However, it could introduce false pseudo-labels, easily leading to corrupting the learning process. This is because classification-based objectives are highly dependent on the accuracy of the given labels~\cite{cross-fault}. For instance, whether decreasing (or increasing) the distance between a sample and a wrong positive (or negative) pseudo-label would depart it from its correct cluster. What's more, iteratively optimizing the classification-based objective would accumulate incorrect information.

In this work, instead of fitting nodes into untrustworthy pseudo-labels, we develop a fault-tolerant way by guiding GNNs to separate two nodes whose positive and negative pseudo-labels target the same class (the opposite pseudo-label). For example, in Figure~\ref{fig:PL_and_PCL}, $C_{1}$ and $C_{2}$ are marked as the opposite pseudo-labels for $\node{1}$ and $\node{2}$ since $\node{1}$'s positive/negative pseudo-label and $\node{2}$'s negative/positive pseudo-label target $C_{1}$/$C_{2}$. Building upon the insights from recently developed Contrastive Learning (CL)~\cite{cpc,infogcl,simclr,velickovic2019deep}, we propose \textbf{Pseudo Contrastive Learning (PCL)} to enhance separation between nodes for GNNs, which can significantly reduce the error probability of PL. To better understand PCL, we take the red area in Figure~\ref{fig:PL_and_PCL} as an example: (a) PPL pulls $\node{1}$ toward $C_{1}$; (b) NPL pushes $\node{1}$ away from $C_{2}$ and $C_{3}$; (c) PCL instead directly pushes $\node{1}$ away from $\node{2}$ via the opposite pseudo labels $C_{1}$ or $C_{2}$. Let $p_{ij}$ (or $\tilde{p}_{ij} = 1-p_{ij}$) represents the probability that $\node{i}$ does (or does not) belong to class $C_{j}$. Table~\ref{tab:explain} further explains that we can decrease the error probability from 0.3 (PPL) and 0.3 (NPL) to 0.18 (our PCL).

\begin{table}[!th]
    \centering
    \caption{Error Probability and Case Explanation of Figure~\ref{fig:PL_and_PCL}}
    \resizebox{1\linewidth}{!}{
    \begin{tabular}{c c c c c c}
    \hline
    \toprule
        \multicolumn{2}{c}{Method} & Error Probability  & Error Case Explanation \\
    \midrule
        PPL 
        & 
        \begin{minipage}[b]{0.15\columnwidth}
		\raisebox{-.4\height}{\includegraphics[width=\linewidth]{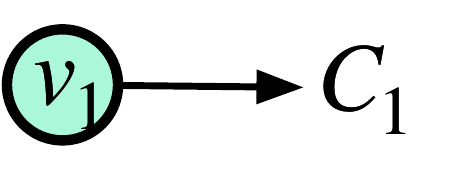}}
		\end{minipage}
		& 
		0.30 = $\tilde{p}_{11}$ 
		& 
		$\node{1}$ dose not belong to $C_{1}$. \\
	 
        NPL 
        & 
        \begin{minipage}[b]{0.15\columnwidth}
		\raisebox{-.4\height}{\includegraphics[width=\linewidth]{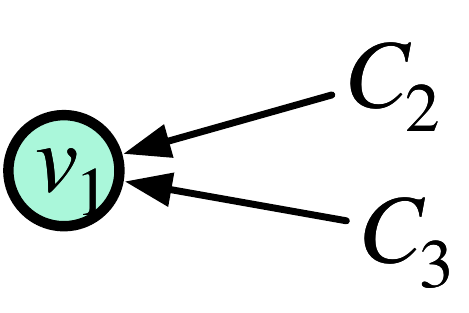}}
		\end{minipage}
		& 
		0.30 = $p_{12} + p_{13}$ 
		& 
		$\node{1}$ belongs to $C_{2}$ or $C_{3}$. \\
%		\makecell{It is the exclusive event of the case \\ that $\node{1}$ does not belong to both $C_{2}$ and $C_{3}$.}  \\
        PCL 
        & 
        \begin{minipage}[b]{0.15\columnwidth}
		\raisebox{-.4\height}{\includegraphics[width=\linewidth]{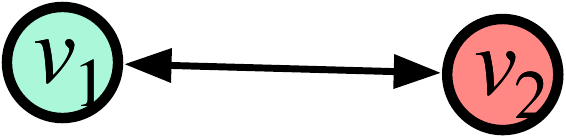}}
		\end{minipage}
		& 
		0.18 = $p_{11}p_{21} + p_{12}p_{22} + p_{13}p_{23}$ 
		& 
		\makecell{$\node{1}$ and $\node{2}$ both \\ belong to the same class.} \\
		
	\bottomrule
    \hline
    \end{tabular}
    }
    
    \label{tab:explain}
\end{table}
% weakness of CL:topological weights
Despite the remarkable success achieved by CL in the graph domain~\cite{velickovic2019deep,hassani2020contrastive,infograph,hdmi,wan2020contrastive,wan2021contrastive,tifa-gcl,sfa}, its generalization to various supervised GNNs remains largely unexplored with two unresolved issues. 

One crucial issue revolves around the traditional CL objective, which relies on carefully selected positive nodes generated through customized augmentation techniques~\cite{dropedge,gca,infogcl} in the unsupervised setting or sampled from nodes of the same class~\cite{wan2020contrastive, wan2021contrastive} in the supervised setting. While strong augmentations can enhance unsupervised CL models by creating "better views" and generating more generalized representations, they might not significantly benefit supervised CL models, as these models require task-specific representations~\cite{simclr}. Additionally, augmentations introduce computational costs and can potentially mislead the message-passing scheme, making it challenging to generalize the approach to any GNN architecture~\cite{tong2021directed}. The supervised CL models often employ the strategy of pulling nodes with the same class (ground-truth or pseudo-labels) closer together. However, optimizing the classification objective (e.g., cross-entropy loss function) could also increase the similarity between training samples of the same class to some extent, demonstrating that the classification objective is capable of maximizing the agreement between positive pairs. Overemphasizing cohesion between positive pairs may not guarantee sufficient separation between negative pairs. The theoretical analysis of \cite{no_pos} also suggests that the positive pairs are relatively drawn close by pushing negative pairs farther away. Furthermore, finding negative nodes, which are more abundant, is comparatively easier than identifying positive nodes. Considering these limitations and inspirations, our proposed PCL focuses on pushing negative pairs farther apart to assist GNNs in learning more separable representations, instead of pulling positive pairs together, which produces trivial improvements but leads to laborious augmentation and positive node sampling.
	
% weakness of CL:augmentation
Another important issue is that the traditional CL objective weights all the negative pairs equally, disregarding their topological relationships. When negative pairs have a smaller topological distance, they become more challenging to separate because a GNN encoder tends to learn similar representations for nearby nodes. To tackle this challenge, it would be beneficial to introduce reasonable biases into different negative pairs based on their topological relationships. Thus, we devise Topologically Weighted Contrastive Loss (TWCL). This loss assigns larger weights to negative pairs with smaller topological distances, allowing PCL to emphasize the separation between nearby negative pairs and further enhance the discriminative ability of GNNs.

Our main contributions are summarized as:

\begin{itemize}
    \item We propose a comprehensive framework called \textbf{Pseudo Contrastive Learning (PCL)} that facilitates Graph Neural Networks (GNNs) in learning discriminative node representations for graph-based SSL. Our novel supervised CL framework eliminates the need for positive pairs, ensuring that GNNs achieve significant separation between nodes.
    
    \item We further enhance the discriminative ability of GNNs by introducing a topologically weighted contrastive loss (TWCL) that incorporates the structural relationships of node pairs into CL. By considering the topological distances between nodes, TWCL provides additional guidance for GNNs to generate more discriminative representations.
   
    \item We enable several commonly used GNN models to achieve considerable performance improvements, surpassing recent state-of-the-art performance in the semi-supervised node classification task. We validate our approach on five real-world graphs, demonstrating the effectiveness and versatility of our proposed PCL framework.
\end{itemize}

\section{Related Work}
\label{sec:related-work}
In this section, we briefly introduce Graph Neural Networks (GNNs) and detail the development of Pseudo Labeling (PL) and Contrastive Learning (CL).

\subsection{Graph Neural Networks}
Motivated by Convolutional Neural Networks (CNNs) in the computer vision domain, many researchers develop GNNs~\cite{gcn,gat,gin,sgc,graphsage,gcnii,appnp,gprgnn} where each central node aggregates information from its neighbors to obtain task-specific node representations. GNNs are typically used to extract low-dimensional node representations from a graph. The learned representations are used to train a strong classifier to address the semi-supervised node classification problem. However, traditional GNNs can only exploit information from limited labeled nodes and their neighbor nodes, ignoring information reserved in massive unlabeled nodes. Therefore, the community shows increasing attention on how to devise a general technique for GNNs to take advantage of more unlabeled nodes. 

\subsection{Pseudo Labeling}

\para{Positive Pseudo Labeling.}\cite{pl} first proposes the idea of PPL which utilizes a trained model to generate pseudo-labels for unlabeled data by the confidence-based thresholding mechanism and then adds them to the training set. Then, several variants of PPL have been proposed and widely used in SSL field~\cite{arazo2020pseudo,pham2019semi,sohn2020fixmatch}. Considering the flexibility and effectiveness of PPL, some works~\cite{li2018deeper,sun2020multi,mutual_pl,dynamic_pl,informative_pl,shgp-yang} attempt to adapt it to GNNs for more improvement in graph-based SSL. The major difference between these works is how to enhance the quality of positive pseudo-labels. \cite{li2018deeper} absorbs PPL into graph-based SSL via combining co-/self-training strategies to find more training nodes in the unlabeled dataset. Similarly,~\cite{sun2020multi} introduces a self-checking mechanism via DeepCluster~\cite{bruna2013spectral} to produce more confident pseudo-labels. \cite{mutual_pl} uses positive pseudo-labels as useful knowledge to teach each peer network and increase the training set. \cite{dynamic_pl} introduces Soft Label Confidence that computes personalized confidence values to label each unlabelled node. To reduce the redundant information carried by high-confidence nodes, \cite{informative_pl} proposes an informativeness measurement to choose reliable nodes. 
 
 Let $\node{i}$ denote node $i$ and $p_{ij}$ denote the element of the prediction matrix $P$. Generally, the positive pseudo-label $y^{+}_{i}$ for $\node{i}$ can be generated as:
 \begin{equation}
 	\label{eq:pl}
	y^{+}_{i} = {\underset{j \in \{1,..,C\}}{\arg\max \, p_{ij}}} \cdot \mathbbm{1}[{\underset{j \in \{1,..,C\}}{\max \, p_{ij} \geq \gamma^{+}}}],
 \end{equation}
 where $\gamma^{+} \in (0,1)$ is a threshold and $C$ is the number of classes. $y^{+}_{i} = 0$ indicates that $\node{i}$ does not have a positive pseudo-label.

\para{Negative Pseudo Labeling.}
 \cite{defense} combines PPL and NPL to select pseudo-labels according to their uncertainties. Then, the positive/negative pseudo-labels are used in classification objective/Negative Learning~\cite{nlnl} to teach the model that $\node{i}$ does/does not belong to its positive/negative pseudo-labels. Let $\aVector{y}{i}^{-} = [y^{-}_{i1},..., y^{-}_{iC}] \subseteq \{0, 1\}^{C}$ be a multi-class binary vector indicating the selected negative pseudo label for $\node{i}$. Then, the generation of $\aVector{y}{i}^{-}$ for $\node{i}$ can be written as:
\begin{equation}
	\label{eq:npl}
	y^{-}_{ic} = \mathbbm{1}[p_{ic} \leq \gamma^{-}],
\end{equation}
where $\gamma^{-} \in (0,1)$ and $c$ is a negative pseudo-label of $\node{i}$ when $y^{-}_{ic} = 1$. 

\subsection{Contrastive Learning}
CL~\cite{cpc,simclr,hjelm2018learning}, known as an instance-discrimination unsupervised method, has demonstrated outstanding performance in representation learning tasks. The generalized CL framework consists of four modules: (1) A stochastic data augmentation module that transforms a given sample from two different views into an anchor sample and a positive sample, which are considered as a positive pair; (2) A negative sampler that selects $K$ negative samples for each anchor sample. An anchor sample and an arbitrary sample from the $K$ negative samples as a negative pair; (3) A shared encoder that extracts representations of each sample; (4) A contrastive objective that maximizes (minimizes) the agreement between the positive (negative) pair(s). The CL method has been extended to graph-structure data and we divide the related works into two categories, i.e.,  unsupervised and semi-supervised Graph Contrastive Learning (GCL).

\begin{figure*}[th!]
    \centering
    \includegraphics[width=1\linewidth]{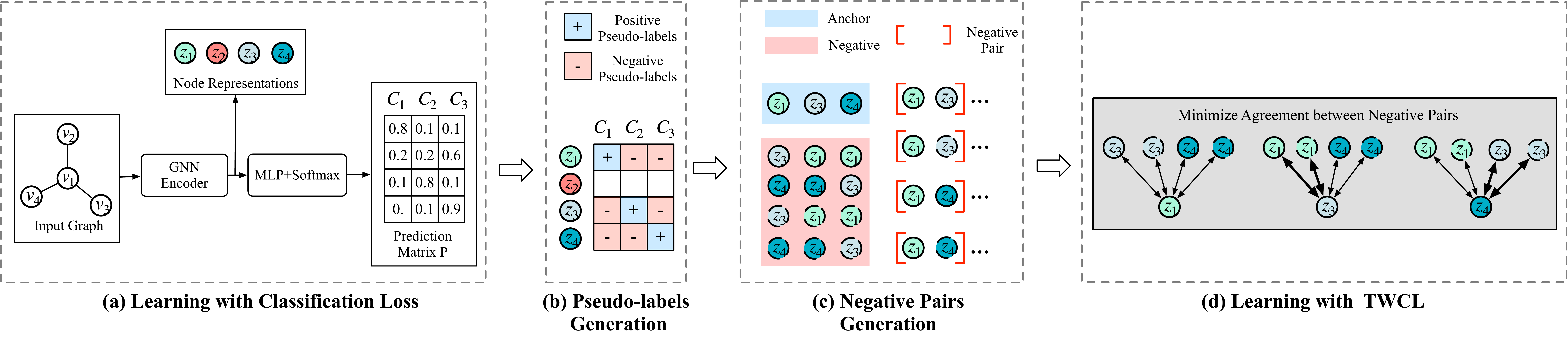}
    \caption{The framework of PCL. In (a), the GNN is trained with classification loss, i.e., cross-entropy loss, to obtain the encoded node representations and prediction matrix $P$. In (b), positive pseudo-labels are generated with $\gamma^{+} = 0.8$, and negative pseudo-labels are produced according to the top-$K$ smallest predicted score in $P$ for each class  ($K = 2$). In (c), nodes with positive pseudo-labels are selected as anchor nodes and those with the opposite pseudo-labels are used as negative nodes. In (d), TWCL minimizes (maximizes) agreement between negative pairs. Thicker lines imply more effort is required to separate the corresponding negative pairs with smaller topological distances.}
    \label{fig:PCL}
\end{figure*}

\para{Unsupervised Graph Contrastive Learning.}
 Recently, numerous works~\cite{velickovic2019deep,hassani2020contrastive,infograph,hdmi,sfa,tifa-gcl,cui2021evaluating,sega,zhang2023spectral,magcl,gcc} have adapted the idea of CL to the graph domain. Unsupervised GCL starts with pre-training a GNN encoder by CL to learn generalizable and task-agnostic representations. Then, it fine-tunes the trained GNN with labels for downstream tasks. Most of these works are devoted to generating appropriate views for graph-structure data via customized data augmentation (e.g., edge dropping) or contrasting modes (e.g., node-node and node-subgraph) for learning better representations.  
  
\para{Semi-supervised Graph Contrastive Learning.}
To learn task-specific representations, semi-supervised GCL~\cite{wan2020contrastive,wan2021contrastive,tifa-gcl,zhang2023dynamic,zhou2023smgcl} combines supervised classification objective and CL objective to train the model in an end-to-end fashion. About the CL part, the difference between these works lies in (1) different views generation (augmentation): \cite{wan2020contrastive} considers the representations from the GCN model and generation model as two views, \cite{wan2021contrastive} regards the prediction from two models as two views, and~\cite{tifa-gcl} augments the graph by adaptively adding or deleting edges; (2) different negative samplers: \cite{wan2020contrastive} samples positive/negative nodes based on the limited labels, \cite{wan2021contrastive} directly regards all the other nodes as negatives, and~\cite{tifa-gcl} chooses negative nodes based on the relative distance between nodes.

\section{Preliminaries}
In this section, we introduce some important notations and the framework of a GNN. 

\para{Notation.}
Let $\mathcal{X}_{L} = \{(\node{i}, \aVector{y}{i})\}_{i=1}^{N_{L}}$ be the labeled node set with $N_{L}$ nodes, where $\aVector{y}{i} \in \mathbb{R}^{C}$ is the one-hot vector indicating $\node{i}$'s label and $C$ is the number of classes. Besides, each node $\node{i}$ has a feature vector $\aVector{x}{i}$. Let $\mathcal{X}_{U} = \{\node{i}\}_{i=N_{L} + 1}^{N_{L} + N_{U}}$ represent the unlabeled node set with $N_{U}$ nodes. We then define a graph $\mathcal{G}=\{\mathcal{X}, \mathcal{E}\}$, where $\mathcal{X} = \mathcal{X}_{L} \bigcup \mathcal{X}_{U}$ is the node set, and $N = N_{L} + N_{U}$ is the number of all nodes. Let $\mathcal{E}=\{e_{ij}\}, i,j\in \{1,2,...,N\}$ denote the edge set, where $e_{ij}$ indicates the edge between $\node{i}$ and $\node{j}$. For simplicity, both node $i$ and its index are denoted by $\node{i}$. Let $[\boldsymbol{q}]_{k}^{-}$ be an operator that extracts $k$ indexes of the top-$k$ smallest values of a vector $\boldsymbol{q} \in \mathbb{R}^{N}$.

\para{GNN.}
Given a $L$-layer GNN model, the $l$-th layer $f^{(l)}$ can be written as:
\begin{equation}
	\aVector{h}{i}^{(l)} = f^{(l)}(\boldsymbol{h}_{i}^{(l-1)}, \{\aVector{h}{j}^{(l-1)} | j \in \mathcal{N}_{i}\}), \nonumber
\end{equation}
where $\mathcal{N}_{i} = \{\node{j} | e_{ij} \in \mathcal{E}\}$ is the neighbor set of $\node{i}$ and $\aVector{h}{i}^{(l)}$ is the latent representations of $\node{i}$ from $l$-th layer ($\aVector{h}{i}^{(0)} = \aVector{x}{i}$). The learned representations of $\node{i}$ is $\aVector{z}{i} = \aVector{h}{i}^{(L)}$. Then, the prediction of $\node{i}$ is 
\begin{equation}
    \aVector{p}{i} = \text{softmax} \left(\text{MLP}(\aVector{z}{i})\right), \nonumber
\end{equation} 
which belongs to the $i$-th row vector of the prediction matrix $P \in \mathbb{R}^{N \times C}$. The MLP layer is adopted to project $\aVector{z}{i}$ onto the prediction space.

\section{Proposed Method: PCL}
In this section, we provide a brief overview of our Pseudo Contrastive Learning (PCL) framework. We begin by presenting a high-level summary of PCL, highlighting its key components and objectives. Next, we delve into the detailed process of generating pseudo-labels and negative pairs. Lastly, we introduce our novel approach called topologically weighted contrastive loss (TWCL), explaining how it enhances the discriminative ability of GNNs by incorporating the structural relationships between node pairs.

\subsection{Overview}
We first introduce four important modules of PCL and present an overview of the training procedure.

\para{Four Modules.} As illustrated in Figure~\ref{fig:PCL}, PCL is composed of four modules as following:

\begin{enumerate}
	\item [(a)] \textbf{Learning with Classification Loss:} 
		We begin by incorporating a standard classification loss function into PCL. By comparing the predictions of a GNN model with the ground-truth labels, the classification loss $\mathcal{L}_{s}$ is computed using the cross-entropy loss between the predicted vector $\aVector{p}{i}$ and the true label vector $\aVector{y}{i}$, for each labeled sample $(\node{i},\aVector{y}{i}) \in \mathcal{X}_{L}$. The formulation of the cross-entropy loss is defined as follows:
		\begin{equation}
		\label{loss:cross-entropy}
			\mathcal{L}_{s} = \sum_{(\node{i},\aVector{y}{i}) \in \mathcal{X}_{L}} \ell_{s}(\aVector{y}{i}, \aVector{p}{i}),
		\end{equation}
		where $\ell_{s}(\aVector{y}{i}, \aVector{p}{i})$ represents the cross-entropy loss function.
		
	\item [(b)] \textbf{Pseudo-labels Generation:}
		To facilitate PCL, we generate pseudo-labels based on the model's predictions. Positive pseudo-labels are obtained by applying a threshold $\gamma^{+}$, while negative pseudo-labels are generated by selecting the top-$K$ smallest predicted scores for each class. More details regarding the pseudo-label generation process can be found in Section~\ref{subsec:pl-generation}.
		
	\item [(c)] \textbf{Negative Pairs Generation:} 
		Based on the predictions obtained from the trained GNN model, PCL incorporates nodes with positive pseudo-labels into the anchor set $\mathcal{A}$. For each class $c$, a negative set ${\mathcal{S}_{c}}$ is constructed by selecting the top-$K$ smallest predicted scores. The process of generating negative pairs is elaborated upon in Section~\ref{subsec:samples_construction}.

	\item [(d)] \textbf{Learning with TWCL:} 
		In this module, we define the topologically weighted contrastive loss, which facilitates learning in PCL. For each anchor node $\aVector{z}{i} \in \mathcal{A}$, its negative pairs are formed by considering all nodes $\aVector{z}{j}$ in the negative set $\mathcal{S}_{y^{+}_{i}}$. The topologically weighted contrastive loss, $\mathcal{L}_{c}$, is then computed as follows:		
		\begin{equation}
		\label{loss:contrastive}
		\mathcal{L}_{c} = \frac{1}{|\mathcal{A}|} \sum_{\aVector{z}{i} \in \mathcal{A}} \ell_{c} (\aVector{z}{i}, \mathcal{S}_{y^{+}_{i}}, \tau),
		\end{equation}
		where $\ell_{c}$ minimizes the agreement between negative pairs. Here, $\tau > 0$ represents a temperature hyperparameter. The definition of $\ell_{c}$ is provided in Section~\ref{subsec:twcl}. 
\end{enumerate}

\para{Overview.}
Firstly, a GNN model is trained on $\mathcal{X}_{L}$ by minimizing Eq.~(\ref{loss:cross-entropy}) with $E_{1}$ epochs. Secondly, PCL generates pseudo-labels based on the prediction from the trained GNN. Thirdly, we select construct negative pairs according to the generated pseudo-labels. Finally, we regard all the nodes as the unlabeled data and train GNN on $\mathcal{X}$ by jointly minimizing classification loss Eq.~(\ref{loss:cross-entropy}) and TWCL Eq.~(\ref{loss:contrastive}) with $E_{2}$ epochs using the following loss:
\begin{equation}
  \mathcal{L} = \mathcal{L}_{s} + \mathcal{L}_{c}.
\end{equation}
It is worth noting that PCL updates all the pairs each iteration after $E_{1}$ epochs to obtain high-quality pairs.

\begin{algorithm}[t]
\caption{Pseudo-labels Generation}
\label{alg:pl-generation}
% \LinesNumbered
\KwIn{Prediction matrix $P$, threshold $\gamma^{+}$, negative node size $K$, number of nodes $N$, number of classes $C$}
\KwOut{ Positive Pseudo-labels $\{y^{+}_{i}\}_{i=1, \cdots, N}$, negative pseudo-labels $\{\aVector{y}{i}^{-}\}_{i=1, \cdots, N}$}
    \For{$i=1$ to $N$}{
                Obtain the positive pseudo-label $y^{+}_{i}$ via Eq.~(\ref{eq:pl}) with the threshold $\gamma^{+}$;\\
        }
    \For{$c=1$ to $C$}{
                \For{$i=1$ to $N$}{
                \If{$i \in [\boldsymbol{P}_{:,c}]_{K}^{-}$}{
                    Choose class $c$ as a negative pseudo-label of $\node{i}$ by setting $y^{-}_{ic} = 1$;
                }
                }
        }
\end{algorithm}

\subsection{Pseudo-labels Generation}
\label{subsec:pl-generation}

In this part, we first introduce the generation of positive and negative pseudo-labels and formalize this process in Algorithm~\ref{alg:pl-generation}.

\para{Positive Pseudo-labels.} 
We adopt the generation method defined by Eq.~(\ref{eq:pl}) with the threshold $\gamma^{+}$ to obtain the positive pseudo-label $y^{+}_{i} \in {1, \cdots, C}$ for each node $\node{i}$. 

\para{Negative Pseudo-labels.}
In contrast to~\cite{defense}, which assigns negative pseudo-labels at the instance level (i.e., the row space of the prediction matrix) based on low predicted scores, we generate negative pseudo-labels at the cluster level using the \textbf{top-$K$ smallest} predicted scores.  Specifically, for a negative pseudo-label vector $\aVector{y}{i}^{-}$, we have $y_{ic}^{-} = 1$ when $i \in [\boldsymbol{P}_{:,c}]_{K}^{-}$, where $\boldsymbol{P}_{:,c}$ is the $c$-th column vector of $P$. $y_{ic}^{-} = 1$ indicates class $c$ is a negative pseudo-label of $\node{i}$. 

Generating negative pseudo-labels at the cluster level using the top-K smallest predicted scores has certain advantages over assigning negative pseudo-labels at the instance level based on low predicted scores. Here are a few reasons why this approach can be considered better:
\begin{enumerate}
	\item [(a)] \textbf{Clearer Separation:} By considering the top-K smallest predicted scores, we are more likely to capture nodes that are dissimilar to the positive pseudo-label nodes, thus maintaining a clearer separation between negative pairs.
	\item [(b)] \textbf{Reduced Ambiguity:} Assigning negative pseudo-labels at the instance level based on low predicted scores can be more ambiguous and prone to noise. Nodes with relatively low scores may still contain valuable information and should not be entirely discarded as negative instances. By focusing on the top-K smallest scores, we can be more confident in selecting nodes that are less likely to be misclassified positive instances, thereby reducing ambiguity in the negative pseudo-label assignment.

\end{enumerate}
 
\begin{figure*}[!th]
    \centering
     \includegraphics[width=0.9\linewidth]{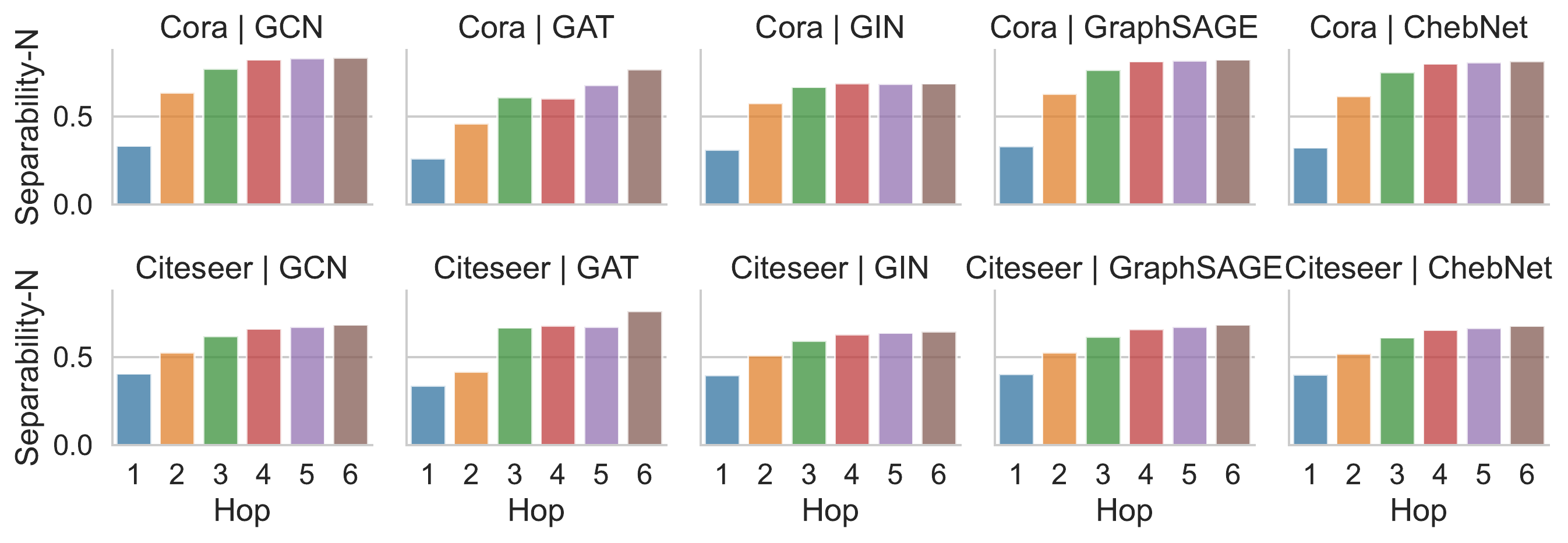}
    \caption{Neighbor separability (Separability-N) illustration on Cora (upper) and Citeseer (bottom). Each bar indicates the mean separability between nodes and their neighbors with different labels from various hops. Larger separability-N implies more separable representations between neighbors. As the topological distance (Hop) decreases/increases, it is harder/easier to obtain separable representations for  GNNs.}
    \label{fig:sep-GNN}
\end{figure*}

\subsection{Negative Pairs Generation}
\label{subsec:samples_construction}

There are two kinds of important nodes, i.e., anchor and negative nodes, used by PCL as follows:
\begin{itemize}
	 \item \textbf{Anchor Nodes}: 
	 	The anchor set $\mathcal{A} = \{\aVector{z}{i} | y_{i}^{+} \neq 0\}$ contains all the nodes with positive pseudo-label.
	 		 			 	
	 \item \textbf{Negative Nodes}: 
	 	For each class $c$, there is a negative set $\mathcal{S}_{c} = \{ \aVector{z}{i} | y_{ic}^{-}=1\}$ that contains $K$ negative nodes. 
\end{itemize}

\begin{algorithm}[t]
\caption{Negative pairs Generation}
\label{alg:pairs-generation}
\KwIn{Positive Pseudo-labels $\{y^{+}_{i}\}_{i=1, \cdots, N}$, negative pseudo-labels $\{\aVector{y}{i}^{-}\}_{i=1, \cdots, N}$, the learned representations $\{\aVector{z}{i}\}_{i=1, \cdots, N}$, number of classes $C$}
\KwOut{Negative Pairs}

\For{$c=1$ to $C$}{
	\For{$i=1$ to $N$}{
		\If{$y^{-}_{ic} = 1$}{
			 Add a negative node $\aVector{z}{i}$ into $\mathcal{S}_{c}$;\\
		}
	}
}

\For{$i=1$ to $N$}{
	\If{$y^{+}_{i} \neq 0$}{
              Add an anchor node $\aVector{z}{i}$ into $\mathcal{A}$;\\
	       \For{$\aVector{z}{j}$ in $\mathcal{S}_{y_{i}^{+}}$}{
                Construct a negative pair $(\aVector{z}{i}, \aVector{z}{j})$;
           }
    }	
}

\end{algorithm}
The whole generation process is described in Algorithm~\ref{alg:pairs-generation}. Furthermore, PCL only uses negative pairs exclusively in the contrastive loss for several reasons:
\begin{itemize}
	\item [(a)] \textbf{Classification Objective Increases Positive Pair Similarity:} The classification objective, such as the cross-entropy loss, naturally encourages the model to increase the similarity between training samples of the same class (positive pairs) to some extent. This means that solely optimizing the classification objective can also enhance the agreement between positive pairs.
	
	\item [(b)] \textbf{Separation Matters More in Supervised Settings:} As we discussed before, cohesion among positive pairs can be readily achieved through the classification objective. Therefore, overemphasizing cohesion may not guarantee sufficient separation between negative pairs. Besides, theoretical analysis, as mentioned in \cite{no_pos}, suggests that pushing negative pairs farther apart is a viable strategy for learning informative representations. By emphasizing the separation of negative pairs, the positive pairs are relatively drawn close. 	
	
	\item [(c)] \textbf{Abundance of Negative Nodes:} In many real-world scenarios, negative nodes tend to be more abundant than positive nodes. It is often easier to find negative nodes to form negative pairs, making them a more practical choice for training. Additionally, identifying positive nodes can be laborious and sometimes uncertain. 
\end{itemize}

\subsection{Topologically Weighted Contrastive Loss}
\label{subsec:twcl}
In this section, we first review the traditional contrastive loss. Then, we empirically demonstrate that GNNs tend to encode indistinguishable representations of node pairs with smaller topological distances compared to larger ones. To obtain more distinguishable representations, we propose topological weighted contrastive loss (TWCL) to focus on the separation between nearby negative pairs.

\para{Traditional Contrastive Loss.}
For each anchor node $\aVector{z}{i} \in \mathcal{A}$, its positive pair is $(\aVector{z}{i}, \aVector{\tilde{z}}{i})$ and $K$ negative pairs are $(\aVector{z}{i}, \aVector{z}{j})_{\aVector{z}{j} \in \mathcal{S}_{y_{i}^{+}}}$. The traditional contrastive loss is to minimize (maximize) the agreement between positive (negative) pairs. We first introduce the contrastive loss defined by a Jensen-Shannon MI estimator~\cite{jsd}. Let $\dis(\aVector{z}{i}, \aVector{z}{j}, \tau) = \sigma{\slfrac{(\text{sim}(\aVector{z}{i}, \aVector{z}{j})}{\tau})}$ denote a discriminator, where $\sigma$ is Sigmoid function and $\text{sim}(\aVector{z}{i}, \aVector{z}{j})$ measures the cosine similarity between $\aVector{z}{i}$ and $\aVector{z}{j}$. For simplicity, we abbreviate the positive contrasting $\dis(\aVector{z}{i}, \aVector{\tilde{z}}{i}, \tau)$ to $\dis^{+}_{ii}$ and the negative contrasting $\dis(\aVector{z}{i}, \aVector{z}{j}, \tau)$ to $\dis_{ij}$. Then, the loss function for $\node{i}$ can be written as:
\begin{small}
	\begin{equation}
    \label{loss:nt-logistic}
    \ell_{c}^{\prime} = \frac{-\left(\ln \dis_{ii}^{+} + \sum_{\aVector{z}{j} \in \mathcal{S}_{y^{+}_{i}}} \ln (1 - \dis_{ij})\right)}{1+K}.
\end{equation}
\end{small}
\para{Poor Separation between Nearby Nodes.}
Given a central node $\node{i}$ and one of its negative node $\node{j}$, the smaller distance between $\node{i}$ and $\node{j}$ makes them more likely to share the similar representations $\aVector{z}{i}$ and $\aVector{z}{j}$ since most GNNs act like a low-pass filter. In Figure~\ref{fig:sep-GNN}, we record the average separability between nodes and their neighbors with different classes from various hops. Then, the $r$-hop neighbor separability is calculated by
\begin{small}
    \begin{equation}
         \text{Separability-N} (r) = \frac{\sum_{i \in \mathcal{X}}\sum_{j \in \mathcal{N}_{i} (r)} \mathbbm{1}_{[\aVector{y}{i} \ne \aVector{y}{j}]} (1 - \text{sim}(\aVector{z}{i},\aVector{z}{j}))}{\sum_{i \in \mathcal{X}} |\mathcal{N}_{i} (r)|}, \nonumber
    \end{equation}
\end{small}
where $\mathcal{N}_{i} (r)$ is the $r$-hop neighbor node set of $\node{i}$ and $|\cdot|$ is the cardinality of set. It demonstrates that nodes that are close to each other are more indistinguishable than nodes that are far apart. As a result, if we consider all the negative pairs to be equally important, it might induce undesired efforts to separate various negative pairs, i.e., negative pairs that are far away from each other are easily separated but those that are close to each other are harder to be separated. Then, it poses a difficulty that close negative nodes cannot be adequately separated from the anchor node using the traditional contrastive loss function. 
	
\para{Topologically Weighted Contrastive Loss.}
To obtain better separation between negative pairs, we incorporate topological information into CL loss and propose topological weighted contrastive loss (TWCL) which puts different weights on various negative pairs based on their structural relationships. Therefore, PCL enforces the model to make more efforts to separate $\node{i}$ away from its negative node $\node{j}$ if they are closer to each other.

To this end, we propose to use Random Walk with Restart (RWR)~\cite{random-walks} which is designed to explore a network's global topology by simulating a particle moving between neighbors iteratively with the probability that is proportional to their edge weights. At each step, the particle starts from a central node $\node{i}$ and randomly moves to its neighbors or returns to $\node{i}$ with probability $q$. As a result, the topological relevance score of $\node{j}$ w.r.t. $\node{i}$ can be measured by the steady probability $\tilde{r}_{ij}$ that the particle will finally stay at $\node{j}$. Then, the $t$-th RWR iteration can be defined as:
\begin{equation}
    \tilde{\boldsymbol{r}}^{(t+1)}_{i} = qA_{rw}\tilde{\boldsymbol{r}}^{(t)}_{i} + (1 - q)\boldsymbol{e}_{i},
\end{equation}
where $A_{rw}$ is the transition probability matrix derived from normalizing the columns of $A$ and $\boldsymbol{e}_{i}$ is the starting vector with $i$-th element $1$ and $0$ for others. The closed solution of $\tilde{\boldsymbol{r}}_{i}$ can be written as:
\begin{equation}
    \aVector{\tilde{r}}{i} = (I - qA_{rw})^{-1}e_{i}.
\end{equation}
The topological relevance score $\aVector{\tilde{r}}{i}$ quantifies the proximity between $\node{i}$ and all the other nodes, reflecting the topological relationships between nodes. Larger $\tilde{r}_{ij}$ indicates closer topological relationship between $\node{i}$ and $\node{j}$. Then, we can obtain TWCL by modifying $\ell_{c}^{\prime}$ in Eq.~(\ref{loss:nt-logistic}) as:
\begin{small}
	\begin{equation}
	\label{loss:TWCL}
   \ell_{c} =- \sum_{\aVector{z}{j} \in \mathcal{S}_{y^{+}_{i}}} w_{ij}\ln (1 - \dis_{ij}),
\end{equation}
\end{small}
where $w_{ij}$ is defined as:
\begin{equation}
	 w_{ij}=
\frac{\exp(\tilde{r}_{ij})}{ \sum{j \in \mathcal{S}_{y^{+}_{i}}} \exp(\tilde{r}_{ij})},\quad i \ne j
\end{equation}
Therefore, $w_{ij}$ tends to stress negative nodes that are located near the central node $\node{i}$.  It is worth noting that $w_{ij}$ can be calculated during the pre-processing stage and stored in the disk so that TWCL will not introduce heavy time overhead. The overall training process of PCL is provided in Algorithm~\ref{alg:pcl}.

\begin{algorithm}[t]
\caption{PCL Training}
\label{alg:pcl}
\KwIn{Feature matrix $X$, adjacency matrix $A$, ground-truth labels $Y$, temperature hyperparameter $\tau$, warmup epochs $E_{1}$, co-training epochs $E_{2}$, topological weights $W$, inputs of Algorithm~\ref{alg:pl-generation} and~\ref{alg:pairs-generation}}
\KwOut{Final prediction of the unlabeled nodes}

\For{$itr=1$ to $E_{1}$}{
    Obtain the prediction of GCN;\\
    Minimize classification loss in Eq.~(\ref{loss:cross-entropy});\\
}
	
\For{$itr=1$ to $E_{2}$}{
        Generate pseudo-labels via Algorithm~\ref{alg:pl-generation};\\
	Generate negative pairs via Algorithm~\ref{alg:pairs-generation};\\	
	Minimize classification loss in Eq.~(\ref{loss:cross-entropy}) and TWCL in Eq.~(\ref{loss:contrastive});\\
}

Obtain the final prediction of unlabeled nodes;\\

\end{algorithm}

\begin{table}[h]
    \centering
    \renewcommand{\arraystretch}{1.5}
    \caption{Datasets Statistics.}
    \resizebox{1\linewidth}{!}{
    \begin{tabular}{l r r r  r r}
    \hline
    \toprule
        Dataset & Nodes & Edges  & Features & Label Rate & Classes\\
    \midrule
        Cora & 2,708 & 5,429  & 1,433 & 5.1\% & 7 \\
        Citeseer & 3,327 & 4,732  & 3,703 & 3.6\% & 6 \\
        Pubmed & 19,717 & 44,338 & 500 & 0.3\%  & 3\\
        Coauthor CS & 18,333 &  81,894 & 6,805 & 1.6\% & 15\\
		Coauthor Physics & 34,493 & 247,962 & 8,415 & 0.2\% & 5 \\
		    \bottomrule
    \hline
    \end{tabular}
}
    
    \label{tab:dataset}
\end{table}

\section{Experiments}
In this section, we evaluate our proposed PCL on the semi-supervised node classification task by applying it to several popular GNNs using five real-world graphs. We also perform comprehensive hyperparameter and ablation studies to better understand our PCL.

\subsection{Datasets}
We use three commonly-used citation networks: Cora, Citeseer, and Pubmed~\cite{cora} and two co-authorship graphs: Coauthor CS and Coauthor Physics~\cite{coauthor}. 

\begin{itemize}

    \item \textbf{Cora}\footnote{https://linqs.soe.ucsc.edu/data} is a citation network between scientific publications that consists of 2708 nodes connected by 5429 edges. The number of classes is 7.
    
    \item \textbf{Citeseer} is a citation network that consists of 3312 scientific publications (nodes) classified into one of 6 classes. There are 4732 edges between all the nodes.
    
    \item \textbf{Pubmed} dataset consists of 44338 links and 19717 scientific publications from PubMed database. Each node is classified into one of three classes.
    
    \item \textbf{Coauthor CS and Coauthor Physics}\footnote{https://www.kdd.in.tum.de/gnn-benchmark} are co-authorship graphs extracted from the Microsoft Academic Graph. Authors are represented as nodes, which are connected by an edge if two authors are both authors of the same paper. 
\end{itemize}
For Cora, Citeseer, and Pubmed, we choose $20$ nodes per class for the training set, $500$ nodes for the validation set, and $1000$ nodes for the test set. For Coauthor CS and Coauthor Physics, we use $20/30$ nodes per class as a training/validation set and the rest nodes as a test set. The statistics of these datasets are described in Table~\ref{tab:dataset}.

\subsection{Experimental Setup}

\para{Hardware\&Software.}
All the experiments were conducted on a single NVIDIA Tesla V$100$ $4096$ Ti with $16$ GB memory size. The CPU is Intel Xeon E$5$-$2650$ v$4$ with $200$ GB memory size. We implement all the models by PyTorch Geometric~\cite{Fey/Lenssen/2019} with the Adam optimizer~\cite{kingma2014adam}.

\para{Backbones.}
We consider six models as backbone models including
\begin{itemize}
    \item \textbf{Multilayer Perceptron (MLP)} that is attribute-based models ignoring the graph structure.
    
    \item \textbf{$L$-localized Graph Convolution Network (ChebNet)}~\cite{chebnet} that uses $L^{th}$-order polynomial in the Laplacian to define the convolution operation on graphs.
    
    \item \textbf{GraphSAGE}~\cite{graphsage} that performs under both the inductive and transductive settings for large graphs.
    
    \item \textbf{Graph Convolution Network (GCN)}~\cite{gcn} that is one of the most popular GNNs. \item \textbf{Graph Attention Network (GAT)}~\cite{gat} that aggregates neighbors' features based on the attention mechanism.
    
    \item \textbf{Graph Isomorphism Network (GIN)}~\cite{gin} that shows equal power to the WL test.
\end{itemize}

\para{Comparing General Techniques.}
We compare PCL against three general techniques including a PL-based method and two discrimination-enhanced methods as follows:
\begin{itemize}
	\item \textbf{UPS}~\cite{defense} that combines PPL and NPL to generate pseudo-labels with uncertainty-aware selection. Lower uncertainty stands for a higher quality of the generated pseudo-label. 
	
    \item \textbf{PairNorm (PN)}~\cite{pairnorm} which is a normalizing scheme to prevent learned embeddings from becoming too similar. 
    
    \item \textbf{DropEdge (DE)}~\cite{dropedge} that removes a certain part of edges during training to generate multiple augmented input graphs.
    
\end{itemize}

\para{Comparing SOTA Graph-based SSL Methods.}
We compare PCL against three PL-based methods with GCN as the backbone model and three semi-supervised GCL methods as follows:
\begin{itemize}
	\item \textbf{Co-training/Self-training/Union/Intersection}~\cite{li2018deeper} that proposes different strategies to select pseudo-labels, i.e., co-training a GCN with a random walk model (Co-training), typical pseudo-label selection (Self-training), and using union or intersection of two pseudo-label sets from Co-training and Self-training. 
	
	\item \textbf{M3S}~\cite{sun2020multi} which selects pseudo-labels with a self-checking mechanism consisting of Deep Cluster algorithm~\cite{deepcluster} and an aligning mechanism.
	
	\item \textbf{InfoGNN}~\cite{informative_pl} which proposes informativeness measure to select high-quality pseudo-labels.
	
	\item \textbf{CG$^{3}$}~\cite{wan2020contrastive} which contrasts node representations from a GNN model and a generation model.

        \item \textbf{CGPN}~\cite{wan2021contrastive} which leverages a contrastive Poisson network to enhance the learning ability of GNNs under limited labeled nodes.
	 
	\item \textbf{TIFA-GCL}~\cite{tifa-gcl} which proposes annotated information to chooses contrasting pairs.
\end{itemize}

\begin{table*}[ht!]
    \caption{Comparison against General Techniques: Classification Accuracy $\pm$ Standard Deviation and Improvement (\%).\label{tab:SSL}}
    \tabcolsep=0.2cm
    \centering
    \renewcommand{\arraystretch}{1.5}
   \resizebox{0.85\linewidth}{!}{
   \begin{tabular}{l l| c c  | c c  | c c | c c | c c}
    \hline
    \toprule
    \multirow{2}{*}{Backbone} & \multirow{2}{*}{Technique} & \multicolumn{2}{c|}{Cora} & \multicolumn{2}{c|}{Citeseer} & \multicolumn{2}{c|}{Pubmed} & \multicolumn{2}{c|}{Coauthor CS} & \multicolumn{2}{c}{Coauthor Physics}\\
    ~ & ~ & Acc. & Imp. & Acc. & Imp. & Acc. & Imp.  & Acc. & Imp.  & Acc. & Imp.\\
    \midrule
        \multirow{5}{*}{MLP} & None & 57.76$\pm$0.78 & 0.00 & 55.33$\pm$1.17 & 0.00 & 72.88$\pm$0.87 & 0.00 & 88.04$\pm$0.83 & 0.00 & 88.78$\pm$1.37 & 0.00\\
        ~ & PN & 60.43$\pm$0.91 & 4.62 & 58.66$\pm$1.42 & 6.01 & 72.39$\pm$1.11 & -0.67 & 88.39$\pm$1.42 & 0.40 & 89.14$\pm$1.14 & 0.41 \\
        ~ & DE & - & - & - & - & - & - & - & - & - & -\\
        ~ & UPS & 64.63$\pm$0.56 & 11.89  & 60.17$\pm$1.97 & 8.75  & 72.59$\pm$0.82 & -0.40  & 88.43$\pm$0.62 & 0.44 & 90.00$\pm$0.97 & 1.37 \\
         
        ~ & \textbf{PCL} & \textbf{66.22$\pm$1.07} & 14.65  & \textbf{65.45$\pm$1.05} & 18.29  & \textbf{73.20$\pm$0.95} & 0.44  & \textbf{88.68$\pm$0.80} & 0.73 & \textbf{90.60$\pm$1.04}	& 1.64 \\
        
        \midrule
        
        \multirow{5}{*}{ChebNet} & None & 77.56$\pm$0.28 & 0.00 & 69.69$\pm$1.01 & 0.00 & 74.68$\pm$0.53 & 0.00 & 89.43$\pm$0.21 & 0.00 & OOM & - \\
        ~ & PN & 77.74$\pm$1.66 & 0.23  & 70.27$\pm$1.85 & 0.83  & 75.85$\pm$0.77 & 1.56  & 89.15$\pm$1.39 & -0.31 & OOM & -\\
        ~ & DE & 78.44$\pm$0.72 & 1.13  & 70.01$\pm$0.52 & 0.46  & 78.03$\pm$0.58 & 4.49  & 89.59$\pm$0.41 & 0.17 & OOM & -\\
        ~ & UPS & 81.78$\pm$0.58 & 5.44  & 70.93$\pm$0.51 & 1.78  & 76.44$\pm$1.02 & 2.36  & 90.47$\pm$0.50 & 1.16 & OOM & -\\
         
        ~ & \textbf{PCL} & \textbf{82.45$\pm$0.86} & 6.30  & \textbf{71.21$\pm$1.34} & 2.18  & \textbf{79.48$\pm$0.52} & 4.46  & \textbf{90.62$\pm$0.65} & 1.33  & OOM & -\\
        
        \midrule
        
        \multirow{5}{*}{GraphSAGE} & None & 75.04$\pm$0.66 & 0.00 & 68.09$\pm$0.81 & 0.00 & 77.30$\pm$0.76 & 0.00 & 91.01$\pm$0.95 & 0.00 & 93.09$\pm$0.48 & 0.00 \\
        ~ & PN & 74.43$\pm$1.14 & -0.81  & 65.22$\pm$1.72 & -4.22  & 76.27$\pm$1.49 & -1.33  & 89.14$\pm$1.13 & -2.05 & 92.41$\pm$1.34 & -0.73 \\
        ~ & DE & 80.66$\pm$0.34 & 7.49  & 69.64$\pm$0.58 & 2.28  & 77.41$\pm$0.39 & 0.14  & \textbf{92.09$\pm$0.48} & 1.19 & 92.73$\pm$1.20 & -0.38\\
        ~ & UPS & 81.83$\pm$0.33 & 9.05  & 70.29$\pm$0.56 & 3.23  & 77.82$\pm$0.56 & 0.67  & 91.35$\pm$0.44 & 0.37 & 93.20$\pm$0.40 & 0.11\\
         
        ~ & \textbf{PCL} & \textbf{82.37$\pm$0.38} & 9.77  & \textbf{71.44$\pm$1.34} & 4.92  & \textbf{79.97$\pm$0.64} & 0.97  & 91.96$\pm$0.88 & 1.04 & \textbf{93.47$\pm$0.21} & 0.40\\
        
        \midrule
        
        \multirow{5}{*}{GCN} & None & 81.57$\pm$0.49 & 0.00 & 70.50$\pm$0.62 & 0.00 & 77.91$\pm$0.37 & 0.00 & 91.24$\pm$0.40 & 0.00 & 92.56$\pm$1.38 & 0.00\\
        ~ & PN & 82.62$\pm$1.14 & 1.29  & 69.98$\pm$1.19 & -0.74  & 78.64$\pm$0.92 & 0.94  & 91.34$\pm$1.94 & 0.11 & 92.78$\pm$1.79 & 0.24 \\
        ~ & DE & 82.02$\pm$0.51 & 0.55  & 71.34$\pm$0.52 & 1.19  & 79.37$\pm$0.54 & 1.87  & 91.41$\pm$0.31 & 0.19 & 92.94$\pm$1.42 & 0.41 \\
        ~ & UPS & 82.35$\pm$0.42 & 0.96  & 72.82$\pm$0.58 & 3.29  & 78.45$\pm$0.37 & 0.69  & 91.62$\pm$0.33 & 0.42 & 93.01$\pm$0.33 & 0.48 \\
         
        ~ & \textbf{PCL} & \textbf{84.28$\pm$0.56} & 3.32  & \textbf{73.60$\pm$0.23} & 4.40  & \textbf{82.32$\pm$0.34} & 3.42  & \textbf{91.81$\pm$0.30} & 0.62 & \textbf{93.47$\pm$0.75} & 0.49 \\
        
        \midrule
        
        \multirow{5}{*}{GAT} & None & 82.04$\pm$0.66 & 0.00 & 71.82$\pm$0.82 & 0.00 & 78.00$\pm$0.71 & 0.00 & 90.52$\pm$0.46 & 0.00 & 91.97$\pm$0.68 & 0.00\\
        ~ & PN & 82.74$\pm$1.18 & 0.85  & 69.09$\pm$0.56 & -3.80  & 77.10$\pm$0.71 & -1.15  & 91.00$\pm$1.00 & 0.53 & 91.46$\pm$0.27 & -0.55 \\
        ~ & DE & 82.63$\pm$0.37 & 0.72  & 71.13$\pm$1.01 & -0.96  & 77.43$\pm$0.56 & -0.73  & 90.74$\pm$0.58 & 0.24 & 92.12$\pm$0.80 & 0.16 \\
        ~ & UPS & 82.17$\pm$0.46 & 0.16  & 72.97$\pm$0.65 & 1.60  & 78.56$\pm$0.87 & 0.72  & 91.26$\pm$0.42 & 0.82 & 92.45$\pm$1.12 & 0.52 \\
         
        ~ & \textbf{PCL} & \textbf{84.48$\pm$0.44} & 2.97  & \textbf{73.22$\pm$0.44} & 1.95  & \textbf{82.14$\pm$0.21} & 2.74  & \textbf{91.74$\pm$0.36} & 1.35 & \textbf{92.65$\pm$0.40} & 0.74 \\
        
        \midrule
        
        \multirow{5}{*}{GIN} & None & 78.48$\pm$0.89 & 0.00 & 66.91$\pm$0.65 & 0.00 & 76.31$\pm$0.78 & 0.00 & 86.93$\pm$1.30 & 0.00 & 88.81$\pm$1.37 & 0.00\\
        ~ & PN & 77.70$\pm$1.44 & -0.99  & 65.20$\pm$1.75 & -2.56  & 77.64$\pm$1.18 & 1.74  & 88.00$\pm$0.83 & 1.23 & 88.02$\pm$1.42 & -0.88\\
        ~ & DE & 78.24$\pm$0.69 & -0.31  & 68.11$\pm$0.93 & 1.79  & 77.59$\pm$1.06 & 1.68  & 87.43$\pm$0.87 & 0.58 & 88.94$\pm$1.31 & 0.14\\
        ~ & UPS & 79.36$\pm$0.39 & 1.12  & 67.55$\pm$0.99 & 0.96  & 76.71$\pm$0.78 & 0.52  & 88.36$\pm$0.36 & 1.65 & 88.68$\pm$2.24 & -0.14\\
         
        ~ & \textbf{PCL} & \textbf{80.81$\pm$0.73} & 2.97  & \textbf{68.91$\pm$1.19} & 2.99  & \textbf{79.87$\pm$0.72} & 2.33  & \textbf{89.70$\pm$0.93} & 3.19 & \textbf{90.37$\pm$0.87} & 1.75\\
    \bottomrule
    \end{tabular}}
\end{table*}

\para{Hyperparameters.} 
For each backbone, we perform a small grid search to decide the best hyperparameters, e.g., dropout rate (dp) in $ \{0.1, \cdots, 0.7\}$, learning rate (lr) in $\{0.01, 0.001, 0.05\}$, hidden dimension (hd) in $\{64, 128, 256\}$, and weight decay (wd) in $\{5e-4, 5e-6, 5e-9\}$.  For each general technique, we directly apply them on the backbone with hyperparameter adjustments according to the suggested configurations from their public source codes. For our PCL, we search the threshold $\gamma^{+} \in \{0.5, \cdots, 0.9\}$ and $K \in \{10, 20, \cdots, 100\}$. Besides, we set the $E_{1} = 200$, $E_{2}=300$, and $\tau=0.05$. 

\subsection{Comparisons}
In this section, we demonstrate the superiority of PCL by analyzing PCL from multiple aspects, i.e., accuracy comparison against general techniques and state-of-the-art (SOTA) PL-based methods, separability comparison, robustness comparison, and augmentation comparison. All experimental results are obtained under $10$ times.

\para{Comparison against General Techniques.}
Table~\ref{tab:SSL} reports the mean accuracies and standard derivations of semi-supervised node classification obtained by all the GNNs w or w/o several general techniques, i.e., PN, DE, UPS, and PCL on Cora, Citeseer, Pubmed, Coauthor CS, and Coauthor Physics. The goal of Table~\ref{tab:SSL} is to examine the contribution of these general techniques to each GNN model, rather than the comparison between different models. Thus, the table should be read by columns for each GNN on each dataset. The best performance for each backbone is highlighted by \textbf{bold}. Besides, we also provide the improvement made by each general technique for the vanilla model. 
\begin{figure*}[!th]
    \centering
     \includegraphics[width=0.85\linewidth]{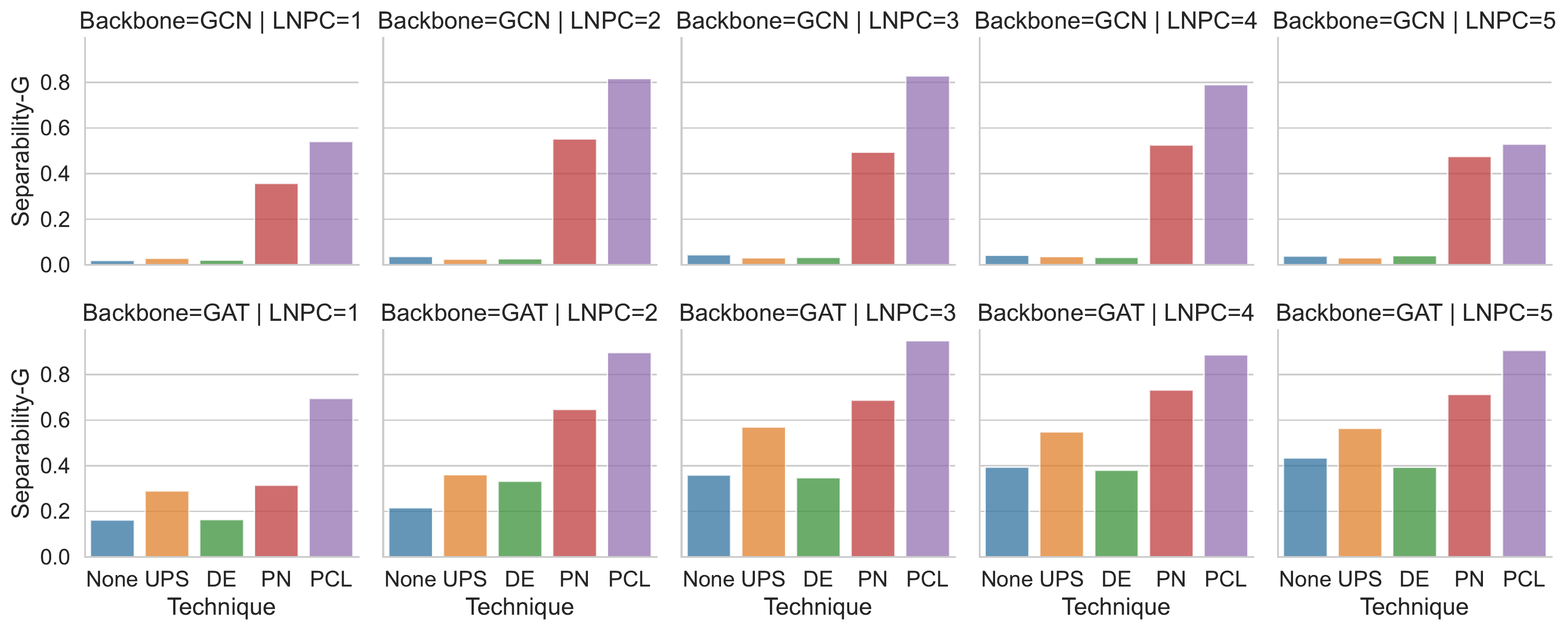}
    \caption{Global separability (Separability-G) comparison against different general techniques, i.e., UPS, DE, PN, and PCL, on Cora. LNPC = $n$ indicates $n$ labeled nodes per class.}
    \label{fig:spe}
\end{figure*}
\begin{table}[h]
    \caption{Comparison against SOTA Graph-based SSL Methods.\label{tab:compar-SOTA}}
    \centering
    \resizebox{0.9\linewidth}{!}
    {
	  \begin{tabular}{l c c c}
	        \hline
	        \toprule
	        Method & Cora & Citeseer & Pubmed\\
	        \midrule
	        Co-training & 79.6 & 64.0 & 77.1 \\
	        Self-training & 80.2 & 67.8 & 76.9 \\
			Union & 80.5 & 65.7 & 78.3 \\
			Intersection & 79.8 & 69.9 & 77.0 \\
			M3S & 78.9 & 69.5 & - \\
			InfoGNN & 82.9 & 73.4 & - \\
			\midrule
			CG$^{3}$ & 83.4$\pm$0.7 & \textbf{73.6$\pm$0.8} & 80.2$\pm$0.8\\
                CGPN & 83.2$\pm$0.4 & 72.7$\pm$0.3 & 79.8$\pm$0.8 \\
			TIFA-GCL & 83.6$\pm$0.8 & 72.5$\pm$0.6 & 79.2$\pm$0.6\\
			\midrule
			\textbf{GCN-PCL} & \textbf{84.28$\pm$0.56}  & \textbf{73.60$\pm$0.23}  & \textbf{82.32$\pm$0.34}\\
	        \bottomrule
	        \hline
	    \end{tabular}
    }
\end{table}Considering MLP does not use graph structure as the input, we do not apply DE to it. We can observe that PCL helps each backbone model to consistently gain better performance on all the graphs. For example, GCN and GAT with PCL can obtain $3.33\%$ and $2.97\%$ increase on Cora. In addition, for MLP that is topology-agnostic, PCL produces a remarkable $14.65\%$ improvement on Cora since our proposed TWCL explicitly incorporates the topological information into optimization. 

\para{Comparison against SOTA Graph-based SSL Methods.}
In Table~\ref{tab:compar-SOTA}, we reuse the public results of Co-training, Self-training, Union, Intersection~\cite{li2018deeper}, CG$^{3}$~\cite{wan2020contrastive}, and TIFA-GCL~\cite{tifa-gcl} and borrow the results of M3S and InfoGNN from~\cite{informative_pl}. All the methods use GCN as the backbone model. We can observe that GCN with PCL (GCN-PCL) obtains better performance in all the cases. For example, in Cora dataset, GCN-PCL achieves $6.8\%$ and $1.7\%$ performance improvements compared to M3S and InfoGNN, which require sophisticated designs to refine the quality of selected pseudo-labels. Even compared with CG$^{3}$ which requires an additional model, i.e., generation model, to generate different views, GCN-PCL can also obtain better classification results.

\begin{figure*}[!th]
    \centering
     \includegraphics[width=0.8\linewidth]{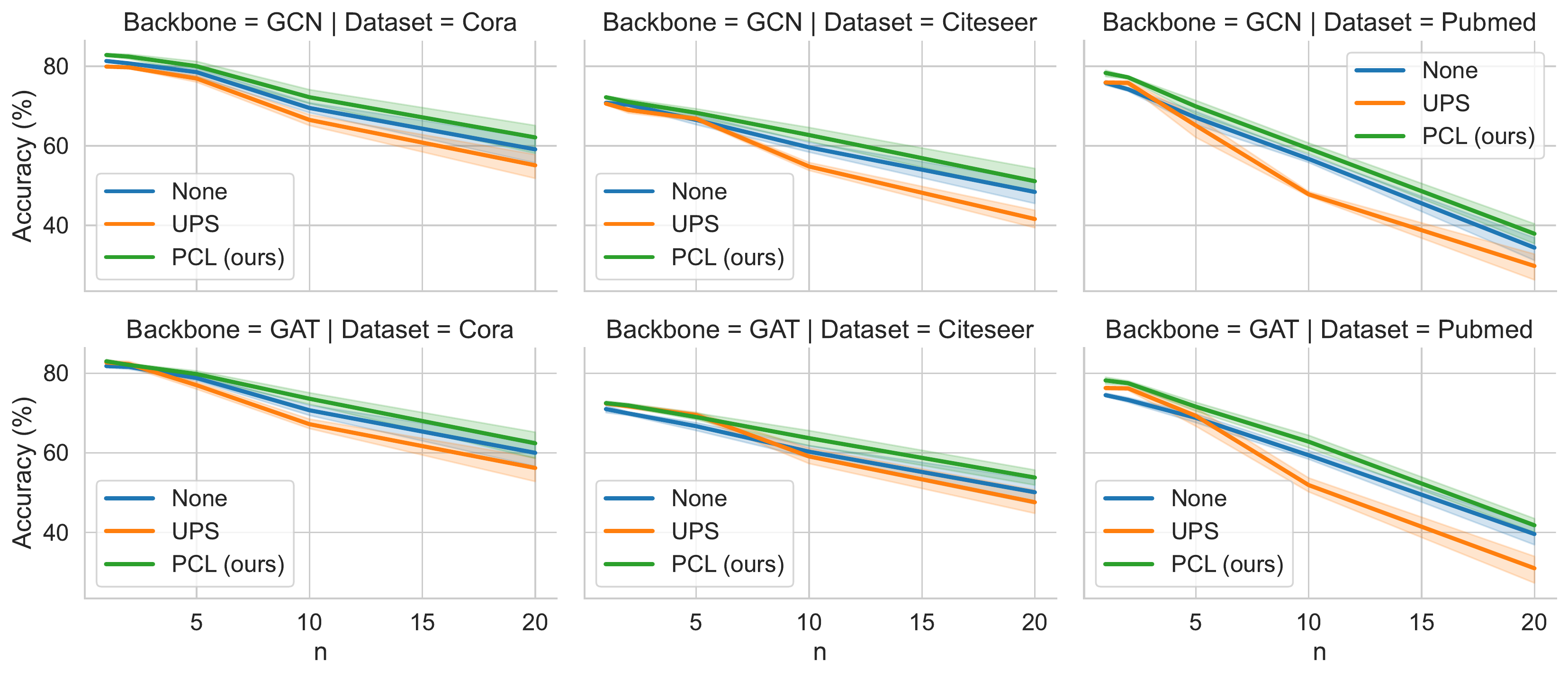}
    \caption{Robustness comparison against UPS with GCN (upper) and GAT (bottom) as the encoders. $n$ indicates that we replace the labels of $n$ randomly selected training nodes with the other $n$ training nodes' (owning different labels). As $n$ increases, UPS even performs worse than the vanilla model since PL might easily introduce noise into training.}
    \label{fig:robustness}
\end{figure*}

\para{Separability Comparison under Scarce Label Setting.}
The number of labeled nodes impacts significantly the discriminative power of GNNs since GNNs naturally learn similar representations within local neighbors. It might result in too similar representations, especially between nodes with different classes, to be classified correctly when labeled nodes decrease. To quantify the global separability of the learned representations, we utilize the mean value of the cosine distance of the final learned representations between each node with all the other ones with different labels. Specifically, the global separability can be calculated by
\begin{equation}
\label{eq:sep}
   \text{Separability-G} = \frac{\sum_{i \in \mathcal{X}}\sum_{j \in \mathcal{X}} \mathbbm{1}_{[\aVector{y}{i} \ne \aVector{y}{j}]} (1 - \text{sim}(\aVector{z}{i},\aVector{z}{j}))}{N(N-1)}. \nonumber
\end{equation}
As shown in Figure~\ref{fig:spe}, we apply UPS/DE/PN/PCL on GCN (upper) and GAT (bottom) using Cora with scarce labeled nodes. Without sufficient labeled nodes, UPS and DE can hardly maintain the separability of representations, while PN can obtain relatively separable representations since it keeps a constant distance between different latent embeddings at each layer. However, it is a class-agnostic technique that fails to maintain inter-class separation. On the contrary, our PCL can help GNNs to learn more separable representations than other counterparts in all cases. 

%\begin{figure*}[!th]
%    \centering
%    \includegraphics[width=0.8\linewidth]{fig_augmentation.pdf}
%    \caption{Ablation study about data augmentation: GCN (upper) and GAT (bottom) equipped with PCL using DropEdge or Dropout (ours). Various ratios $\in \{0.1, \cdots, 0.9\}$ represent the edge sampling rate or Dropout rate.}
%    \label{fig:augmentation}
%\end{figure*}

\para{Robustness Comparison.}
PL enjoys pleasing versatility but suffers from poor robustness mainly induced by the noise from incorrect pseudo-labels. In Figure~\ref{fig:robustness}, we first randomly select $n$ training nodes and then replace their labels with the other $n$ training nodes that belong to the different classes from the selected nodes. We evaluate the robustness of UPS and PCL with GCN and GAT as the encoders using three graphs, on the noisy training set with $n \in \{1, 2, 5, 10, 20\}$ randomly selected training nodes. We can observe a faster performance decline of UPS than the vanilla model as more noisy training nodes are introduced ($n$ increases). This is because UPS treats the generated (possibly incorrect) pseudo-labels as the ground truths for training, thus resulting in unsatisfied performance. In contrast, our PCL is more fault-tolerant by pushing away nodes instead of assigning specific but untrustworthy category labels. 

\begin{table}[!h]
    \centering
    \caption{Default Setting}
    \resizebox{0.9\linewidth}{!}{
    \begin{tabular}{r r r r r r r r r r r}
        \hline
        \toprule
        $\gamma^{+}$ & $K$ & $\tau$ & $E_{1}$ & $E_{2}$ & hd & wd & lr & dp\\
        \midrule
        0.5  & 20 & 0.05 & 200 & 300 & 64 & 5e-4 & 0.01 & 0.5\\
        \bottomrule
        \hline
    \end{tabular}
    }
    \label{tab:def-setting}
\end{table}

\subsection{Hyperparameters Studies}
We evaluate the influences of two important hyperparameters, i.e., threshold $\gamma^{+}$ and negative sample size $K$. All the experiments are conducted $10$ times under the default setting in Table~\ref{tab:def-setting}.
\begin{figure}[!h]
    \centering
     \subfigure{\includegraphics[height=3.5cm, width=4.2cm]{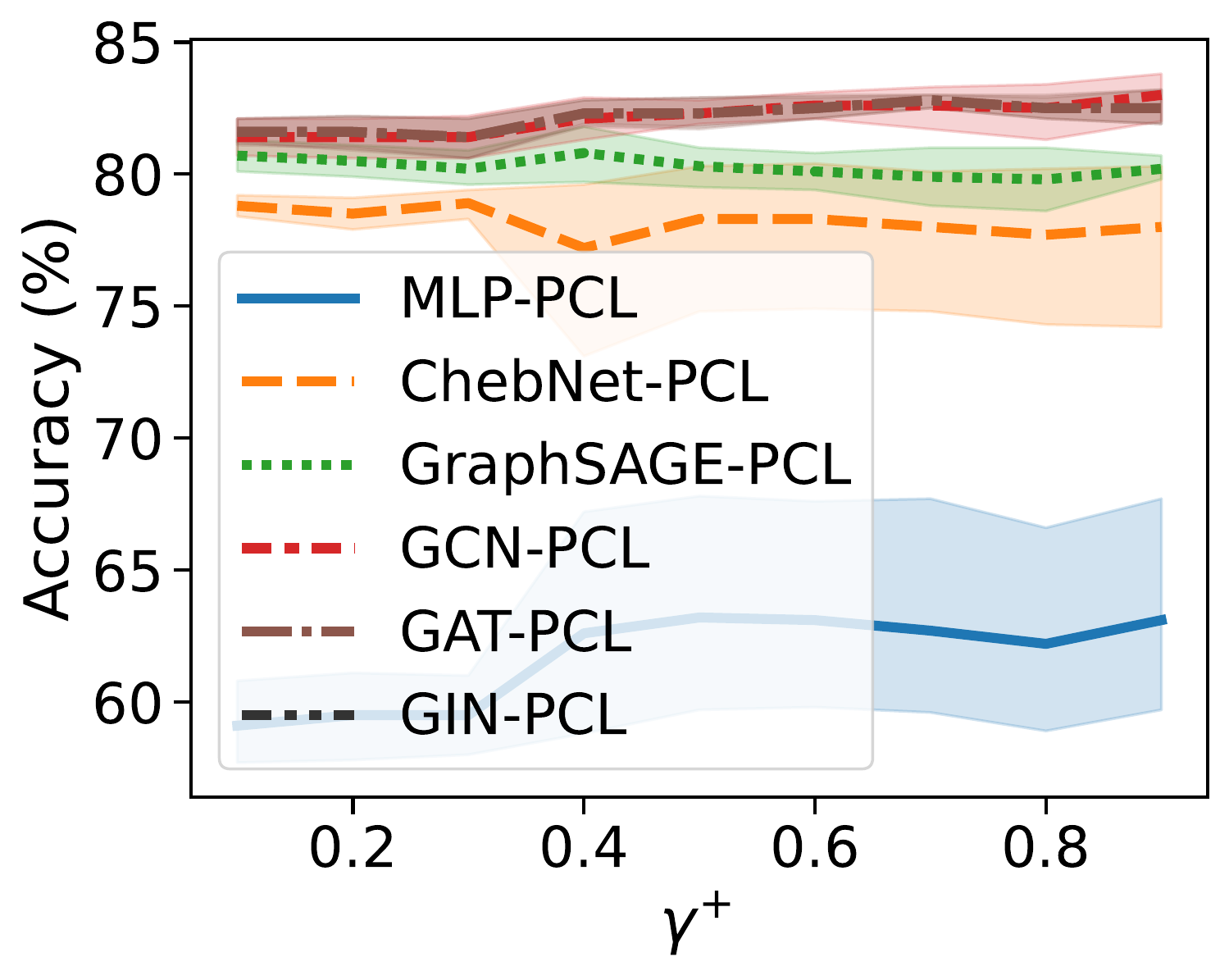}}
     \hfil
     \subfigure{\includegraphics[height=3.5cm, width=4.2cm]{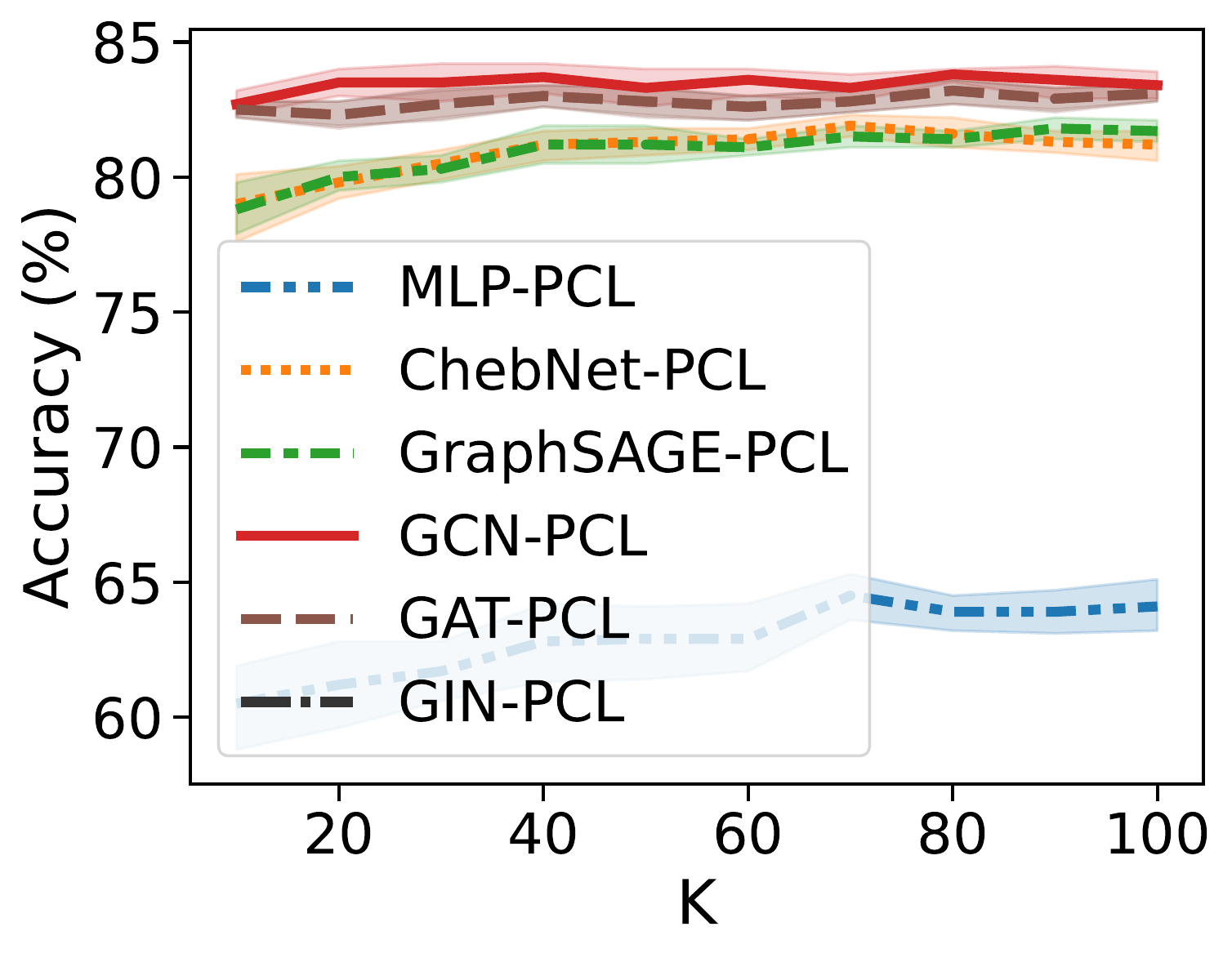}}
    \caption{Hyperparameter analysis about threshold $\gamma^{+}$ (left) and negative nodes size $K$ (right).}
    \label{fig:hyper}
\end{figure}

\para{Threshold $\gamma^{+}$.}
It is the hyperparameter that (1) possibly adjusts the size of the anchor set $\mathcal{A}$ since confident samples tend to be rare, and (2) regulates the negative impact of the potential noise on training. Figure~\ref{fig:hyper} (left) shows that higher $\gamma^{+}$ often guarantees better performance. Besides, larger $\gamma^{+}$ results in higher-quality and fewer anchor nodes which require a less computational cost.

\para{Negative Nodes Size $K$.}
In Figure~\ref{fig:hyper} (right), we visualize the changing trend of accuracy w.r.t varying negative nodes size $K$. We can observe that GNNs with PCL can benefit from increasing $K$ in the interval of 10 to 60. Besides, MLP-PCL always yields performance improvement as $K$ increases. The negative nodes that PCL provides for the contrastive task are always the $K$ nodes which most unlikely belong to the same label as the anchor node. Therefore, the whole training can benefit most from our selected negative nodes.

\subsection{Ablation Studies}
In this section, we conduct an ablation study to analyze the components of PCL and their impact on the performance of the backbone model. Specifically, we investigate the following aspects and present the overall results in Table~\ref{tab:ablation}.

%\begin{figure}[!h]
%    \centering
%    \includegraphics[width=0.95\linewidth]{fig_TWCL}
%    \caption{Ablation study about TWCL: MLP (upper) and GCN (bottom) equipped with PCL w or w/o TWCL using $K \in \{2, 5, 10, 15, 20\}$ negative nodes.}
%    \label{fig:abl_twcl}
%\end{figure}
\begin{table*}[!h]
    \centering
    \caption{Ablation Studies}
    \begin{tabular}{ll cccccc}
        \hline
        \toprule
        \multicolumn{2}{c}{Component}  & Cora & Citeseer & Pubmed & Coauthor CS & Coauthor Physics\\
        \midrule
         \multicolumn{2}{c}{UCL} & 83.30$\pm$0.74 & 72.98$\pm$0.41 & 79.82$\pm$0.71 & 91.47$\pm$0.42 & 93.12$\pm$0.58\\
         \multicolumn{2}{c}{TWCL (ours)} & \textbf{84.28$\pm$0.56} & \textbf{73.60$\pm$0.23}  & \textbf{80.58$\pm$0.52} & \textbf{91.81$\pm$0.30} & \textbf{93.47$\pm$0.75}\\
        \midrule
        
		\multirow{2}{*}{PPL} & w P & 83.03$\pm$0.57 & 71.47$\pm$0.54 & 80.32$\pm$0.55 & 91.56$\pm$0.58 & 92.79$\pm$1.07\\
		~ & w/o P & 82.70$\pm$0.74 & 71.36$\pm$0.71 & 80.26$\pm$0.93 & 91.53$\pm$0.87 & 93.00$\pm$0.56\\
		
		\cmidrule(lr){2-7}
		
		\multirow{2}{*}{NPL} & w P & 82.65$\pm$1.04 & 71.37$\pm$0.65 & 78.58$\pm$0.92 & 90.72$\pm$0.79 & 92.77$\pm$0.98\\
		~ & w/o P & 81.79$\pm$1.05 & 71.29$\pm$0.68 & 78.90$\pm$0.61 & 91.05$\pm$0.60 & 92.74$\pm$0.89\\
		
		\cmidrule(lr){2-7}
		
		\multirow{2}{*}{Top-K (ours)} & w P & 83.69$\pm$0.44 & 72.11$\pm$0.47 & 80.49$\pm$0.31 & 91.43$\pm$0.33 & 93.21$\pm$0.59\\
		~ & w/o P & \textbf{84.28$\pm$0.56} & \textbf{73.60$\pm$0.23}  & \textbf{80.58$\pm$0.52} & \textbf{91.81$\pm$0.30} & \textbf{93.47$\pm$0.75}\\
        \bottomrule
        \hline
    \end{tabular}
    \label{tab:ablation}
\end{table*}

\para{Loss Function Variants.}
We compare the Unweighted Contrastive Loss (UCL) with our proposed Topologically Weighted Contrastive Loss (TWCL). The goal is to assess the effectiveness of TWCL in enhancing the discriminative ability of the model. We observe that TWCL consistently outperforms UCL across all datasets, indicating that the incorporation of topological information in the contrastive loss function is effective in improving the model's performance.

\para{Negative Sample Sampling Strategies.}
We explore different strategies for selecting negative samples. We compare our top-$K$ negative sampling against PPL and NPL variants, each with or without positive pairs. The top-K negative sampling strategy (ours) consistently outperforms other strategies, demonstrating its effectiveness in selecting informative negative samples. Therefore, the whole training benefits most from our proposed Top-$K$ negative sampling strategy because PCL always feeds the contrastive task with $K$ negative nodes which most unlikely belong to the same label as the anchor node. 

\para{Positive Pairs Utilization.}
We examine the use of positive pairs in the PCL framework. Two variants are considered: one with positive pairs (w P) and one without positive pairs (w/o P). The positive pairs are generated based on the positive pseudo-labels. That is, node $j$ becomes a positive nodes for node $i$ if $y^{+}_{i} = y^{+}_{j}$.  We aim to understand the role of positive pairs in PCL. The experiments with and without positive pairs show that the absence of positive pairs (w/o P) does not lead to a significant drop (for PPL and NPL) in performance and even achieves performance improvements in some cases. This suggests that emphasizing negative pairs is effective enough in semi-supervised settings. The reason why using positive pairs in PPL and NPL can lead to improvements while not in PCL's top-$K$ Negative Sampling lies in the nature of the negative sample selection process and the potential introduction of noise. In PPL and NPL, negative sample selection is based on pseudo-labels and a confidence threshold. It might result in nodes that are not truly negative nodes being included because they may have been incorrectly assigned pseudo-labels. PCL adopts the top-$K$ Negative Sampling strategy, which ensures the selection of higher-quality negative nodes. As a result, PCL's negative nodes are of higher quality and less likely to introduce noise, which is why it does not require the use of positive pairs to improve performance.

\section{Conclusion}
In this paper, we propose a fault-tolerant way to help GNNs obtain improvement from a model prediction by constructing pseudo-contrasting pairs instead of pseudo-labels. In particular, we devise a simple yet effective method, PCL, which acts like a representation regularizer benefiting graph-based SSL. The extensive experiments reveal the effectiveness of our method and sufficient ablation studies explain the significance of each design. We expect to generalize our method to other fields, i.e., computer vision and natural language processing, in the future. 

 \section*{Acknowledgments}
This work was supported in part by the National Natural Science Foundation of China under Grants 62133012, 61936006, 62073255, and 62303366, in part by the Innovation Capability Support Program of Shaanxi under Grant 2021TD-05, and in part by the Key Research and Development Program of Shaanxi under Grant 2020ZDLGY04-07.

\bibliographystyle{elsarticle-num}
\bibliography{refs}


%\begin{thebibliography}{00}
%%% Text of bibliographic item
%
%\bibitem{}
%
%\end{thebibliography}

\end{document}